\documentclass[letterpaper, 10 pt, conference]{ieeeconf}  

\IEEEoverridecommandlockouts                              

\usepackage[hidelinks]{hyperref}
\usepackage{graphicx}
\usepackage[export]{adjustbox}
\usepackage{float}
\usepackage{pgf}
\usepackage{pgfplots}
\usepackage{amsmath}
\usepackage[font=footnotesize]{caption}
\usepackage[font=footnotesize]{subcaption}
\usepackage{mwe}

\newcommand\copyrighttext{%
	\footnotesize \textcopyright 2020 IEEE.  Personal use of this material is permitted.  Permission from IEEE must be obtained for all other uses, in any current or future media, including reprinting/republishing this material for advertising or promotional purposes, creating new collective works, for resale or redistribution to servers or lists, or reuse of any copyrighted component of this work in other works.\\
	DOI: 10.1109/RoboSoft48309.2020.9116060}

\newcommand\copyrightnotice{%
	\begin{tikzpicture}[remember picture,overlay]
	\node[anchor=north,yshift=-10pt] at (current page.north) {\fbox{\parbox{\dimexpr\textwidth-\fboxsep-\fboxrule\relax}{\copyrighttext}}};
	\end{tikzpicture}%
}

\title{\LARGE \bf
Towards vision-based robotic skins:\\ a data-driven, multi-camera tactile sensor
}

\author{Camill Trueeb, Carmelo Sferrazza and Raffaello D'Andrea$^{1}$
\thanks{$^{1}$The authors are members of the Institute for Dynamic Systems and Control, ETH Zurich, Switzerland. Email correspondence to Carmelo Sferrazza {\tt\small csferrazza@ethz.ch}}%
}

\begin{document}

\maketitle
\thispagestyle{empty}
\pagestyle{empty}

\copyrightnotice

\begin{abstract}

This paper describes the design of a multi-camera optical tactile sensor that provides information about the contact force distribution applied to its soft surface. This information is contained in the motion of spherical particles spread within the surface, which deforms when subject to force. The small embedded cameras capture images of the different particle patterns that are then mapped to the three-dimensional contact force distribution through a machine learning architecture. The design proposed in this paper exhibits a larger contact surface and a thinner structure than most of the existing camera-based tactile sensors, without the use of additional reflecting components such as mirrors. A modular implementation of the learning architecture is discussed that facilitates the scalability to larger surfaces such as robotic skins.

\end{abstract}
\graphicspath{{./pngs/},{./pdfs/},{./pgfs/}}

\section{INTRODUCTION} \label{sec:introduction}
Research in whole-body tactile sensing \cite{wholebody_book} aims to provide robots with the capability of fully exploiting contact with objects to perform a wide range of tasks. As an example, humans often use both their hands and arms to transport large and heavy boxes, exploiting the feedback from their tactile receptors and the compliance of their soft skin.

The recent advances in computer vision and machine learning have drawn increasing attention towards vision-based tactile sensors, often referred to as optical tactile sensors. These sensors generally employ an optical device to provide high-resolution information about the deformation of their soft surface when subject to external forces. As an example, the motion of spherical particles embedded within a soft, transparent gel is captured by an RGB camera in \cite{sferrazza_sensors} to render feedback about the force distribution that causes the gel's deformation. A typical drawback of the camera-based approaches is the bulkiness of their main sensing unit. Moreover, the minimum focal distance of commercial cameras usually implies the need for additional space between the camera lens and the soft gel, in which the monitored patterns are embedded, e.g., markers, particles, etc., leading to an additional increase of the overall size. Even in the case of close-focus lenses, placing the soft surface too close to the camera generally leads to a reduced field of view (FOV).

\begin{figure}[th]
\centering
\includegraphics[width=1\columnwidth,left]{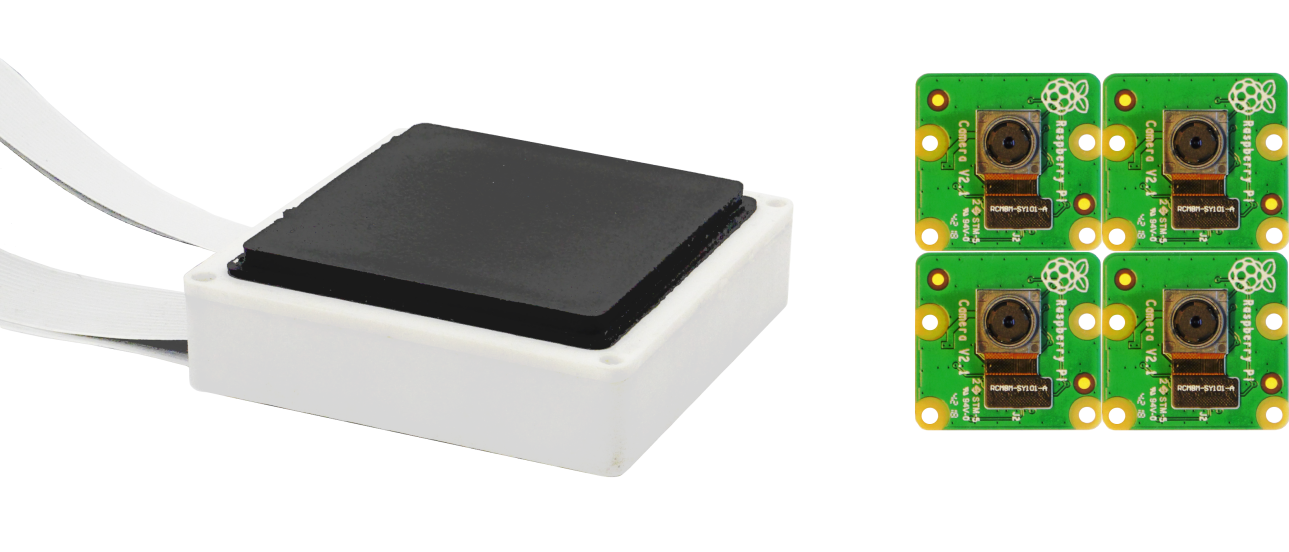}
\caption{The tactile sensor presented in this article has a reduced thickness compared to most of the camera-based tactile sensors in the literature. In this figure, it is shown next to the four embedded camera modules placed below the soft sensor's surface, which measures 49$\times$51 mm. }
\label{fig:sensorpic}
\end{figure}
This paper proposes a multi-camera design to tackle the issues mentioned, leading to a relatively thin overall structure (about 17.5 mm, see Fig. \ref{fig:sensorpic}) and retaining the scalability to larger surfaces. Four embedded cameras are equipped with close-focus lenses and are placed next to each other to cover an increased FOV. A deep neural network (DNN) is trained to reconstruct the three-dimensional contact force distribution applied to the sensor's surface, directly processing the pixel intensities captured on the images. The architecture employed here exhibits a modular structure to increase the software scalability for the implementation on larger surfaces or in the case of a single camera replacement. In fact, a generalization experiment is performed by training the DNN on a subset of the available cameras. A considerably smaller part of the network is then retrained once a new camera is added, resulting in shorter training times and lower data requirements, while generalizing to the entire surface.

The DNN is deployed in real-time, leveraging the capabilities of a state-of-the-art System-on-Module, provided with an integrated GPU. The resulting sensing pipeline predicts the contact force distribution 40 times per second.

\subsection{Related work}
Several physical principles have been applied with the objective of providing robots with an equivalent of the human sense of touch. In fact, a number of categories of tactile sensors exist in the literature, e.g., resistive \cite{resistive_ts}, piezoelectric \cite{piezoelectric_ts} and capacitive \cite{capacitive_ts}. A survey of the different categories is provided in \cite{survey_ts}. Similarly, various examples of tactile skins using the different sensing principles and scalable to large surfaces have been described, see for example \cite{fabric_skin,capacitive_skin}.

Vision-based tactile sensors are based on optical devices, which track visual features related to the deformation of a soft surface. Beside RGB cameras, depth cameras \cite{softbubble} and dynamic vision sensors \cite{dvs_sensor} have been employed in a similar manner. Optical tactile sensors show high resolution, ease of manufacture and low cost, despite a larger thickness compared to the other categories. For an overview of the different types of optical tactile sensors, see \cite{kazu_survey, yamaguchi_survey}.

In \cite{yamaguchi_skin}, the viability of an optical tactile skin is discussed. The availability of inexpensive and low power GPUs is indicated as a possible solution to enable the real-time processing of a large number of tactile images. Two cameras were mounted on each finger of a soft robotic gripper in \cite{exoskeleton_finger} to classify the shape and size of an object. The classification is performed by means of a DNN that takes as input the concatenation of the two images. A finger-shaped gripper is presented in \cite{gelslim}. Tactile imprints are redirected via a mirror towards a camera to increase the sensor compactness. Two cameras are used in \cite{muscularis} to reconstruct the 3D displacement of inner markers in a soft tactile muscularis.

In order to overcome the complexity of interpreting the tactile information, several learning-based approaches have been applied to measure various tactile quantities. The location and depth of an indentation are reconstructed in \cite{overlapping_signals} on a sensor based on an array of light emitters and receivers. In \cite{gelsight_sensors} a deep learning architecture estimates the total force and torque applied to a tactile sensor, which uses photometric stereo and markers painted on its surface. In \cite{fem_access}, a neural network reconstructs the contact force distribution applied to the soft surface of a vision-based sensor. Ground truth labels are provided via the use of simulations based on the finite element method (FEM). In order to share the knowledge acquired from data across different sensors, a transfer learning approach is proposed in \cite{transfer_learning_sferrazza}.

The approach presented here is based on four cameras placed at a short distance from the observed surface, which has a random spread of spherical particles embedded. The choice of the components and the data-driven approach make it possible to obtain a thin structure without the use of additional reflecting components, hence simplifying manufacture. The network architecture employed is tailored to the use of multiple cameras, introducing modularity features and facilitating the scalability of the approach. The resulting pipeline reconstructs with high accuracy the contact force distribution applied by a sample indenter to the soft surface of the sensor, including the regions where the indentation is not fully covered by the FOV of a single camera.

\subsection{Outline}
The sensing principle and the hardware used for the experiments are described in Section \ref{sec:sensor_design}. A dimensioning analysis then discusses the possibility of further reducing the thickness of the sensor. The data collection and the learning architecture are detailed in Section \ref{sec:method}. The results and a modularity evaluation are presented in Section \ref{sec:results}. Remarks in Section \ref{sec:conclusion} conclude the paper.

\section{SENSOR DESIGN} \label{sec:sensor_design}
A four-camera design is first introduced in this section. Compared to previous work it shows a thinner sensor with a larger sensing surface. The outlook for a further reduction of the thickness of the design is discussed in the second part of the section.

\subsection{Hardware}

\begin{figure}[h]
    \centering
    \begin{subfigure}[b]{0.35\columnwidth}
        \centering
\begingroup%
  \makeatletter%
  \providecommand\color[2][]{%
    \errmessage{(Inkscape) Color is used for the text in Inkscape, but the package 'color.sty' is not loaded}%
    \renewcommand\color[2][]{}%
  }%
  \providecommand\transparent[1]{%
    \errmessage{(Inkscape) Transparency is used (non-zero) for the text in Inkscape, but the package 'transparent.sty' is not loaded}%
    \renewcommand\transparent[1]{}%
  }%
  \providecommand\rotatebox[2]{#2}%
  \newcommand*\fsize{\dimexpr\f@size pt\relax}%
  \newcommand*\lineheight[1]{\fontsize{\fsize}{#1\fsize}\selectfont}%
  \ifx\svgwidth\undefined%
    \setlength{\unitlength}{113.38582677bp}%
    \ifx\svgscale\undefined%
      \relax%
    \else%
      \setlength{\unitlength}{\unitlength * \real{\svgscale}}%
    \fi%
  \else%
    \setlength{\unitlength}{\svgwidth}%
  \fi%
  \global\let\svgwidth\undefined%
  \global\let\svgscale\undefined%
  \makeatother%
  \begin{picture}(1,0.85)%
    \lineheight{1}%
    \setlength\tabcolsep{0pt}%
    \put(0,0){\includegraphics[width=\unitlength,page=1]{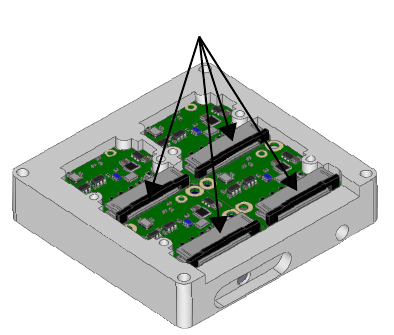}}%
    \put(0.12513859,0.79220505){\color[rgb]{0,0,0}\makebox(0,0)[lt]{\lineheight{1.25}\smash{\begin{tabular}[t]{l}Flat Ribbon Connectors\end{tabular}}}}%
  \end{picture}%
\endgroup%

        \caption{bottom view}    
        \label{fig:bottom}
    \end{subfigure}
    \hspace{20pt}
    \begin{subfigure}[b]{0.35\columnwidth}  
        \centering 
\begingroup%
  \makeatletter%
  \providecommand\color[2][]{%
    \errmessage{(Inkscape) Color is used for the text in Inkscape, but the package 'color.sty' is not loaded}%
    \renewcommand\color[2][]{}%
  }%
  \providecommand\transparent[1]{%
    \errmessage{(Inkscape) Transparency is used (non-zero) for the text in Inkscape, but the package 'transparent.sty' is not loaded}%
    \renewcommand\transparent[1]{}%
  }%
  \providecommand\rotatebox[2]{#2}%
  \newcommand*\fsize{\dimexpr\f@size pt\relax}%
  \newcommand*\lineheight[1]{\fontsize{\fsize}{#1\fsize}\selectfont}%
  \ifx\svgwidth\undefined%
    \setlength{\unitlength}{113.38582677bp}%
    \ifx\svgscale\undefined%
      \relax%
    \else%
      \setlength{\unitlength}{\unitlength * \real{\svgscale}}%
    \fi%
  \else%
    \setlength{\unitlength}{\svgwidth}%
  \fi%
  \global\let\svgwidth\undefined%
  \global\let\svgscale\undefined%
  \makeatother%
  \begin{picture}(1,0.85)%
    \lineheight{1}%
    \setlength\tabcolsep{0pt}%
    \put(0,0){\includegraphics[width=\unitlength,page=1]{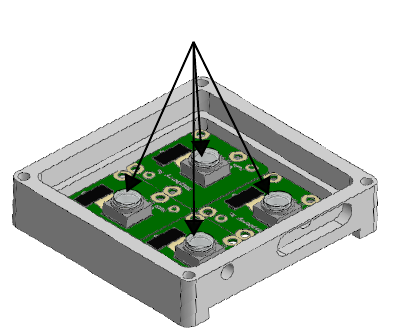}}%
    \put(0.37607548,0.77372284){\color[rgb]{0,0,0}\makebox(0,0)[lt]{\lineheight{1.25}\smash{\begin{tabular}[t]{l}Cameras\end{tabular}}}}%
  \end{picture}%
\endgroup%

        \caption{top view} 
        \label{fig:top}
        \end{subfigure}
    \caption{The sensor's base structure accommodates four Raspberry Pi v2 camera interface boards with flat ribbon cable connectors (a) and the cameras mounted on top (b).}
    \label{fig:topbottom}
    \end{figure}

The optical tactile sensor is based on the tracking of unicolored particles (polyethylene  microspheres with  a  diameter  of  150  to  180$\mu$m) randomly spread within a soft, transparent silicone gel. The motion of the particles is captured by four rectangularly arranged cameras (Raspberry Pi Camera Module v2), see Fig. \ref{fig:topbottom}. These cameras capture 40 frames per second at a resolution of 320$\times$240 pixels. The frames are eventually cropped and downsampled to 128$\times$128 pixels. In order to reduce the thickness of the sensor, the default Raspberry Pi camera lenses are replaced by fisheye lenses originally mounted on Sincerefirst SF-C7251OV-H133 cameras. The lenses are mounted onto the camera frames over distance rings, whose thickness is designed to obtain the desired focus. Finally, an LED board is placed over the camera array to provide uniform brightness.

\begin{figure}[!htb]
\hspace{-40pt}
\includegraphics[width=1.3\columnwidth,left]{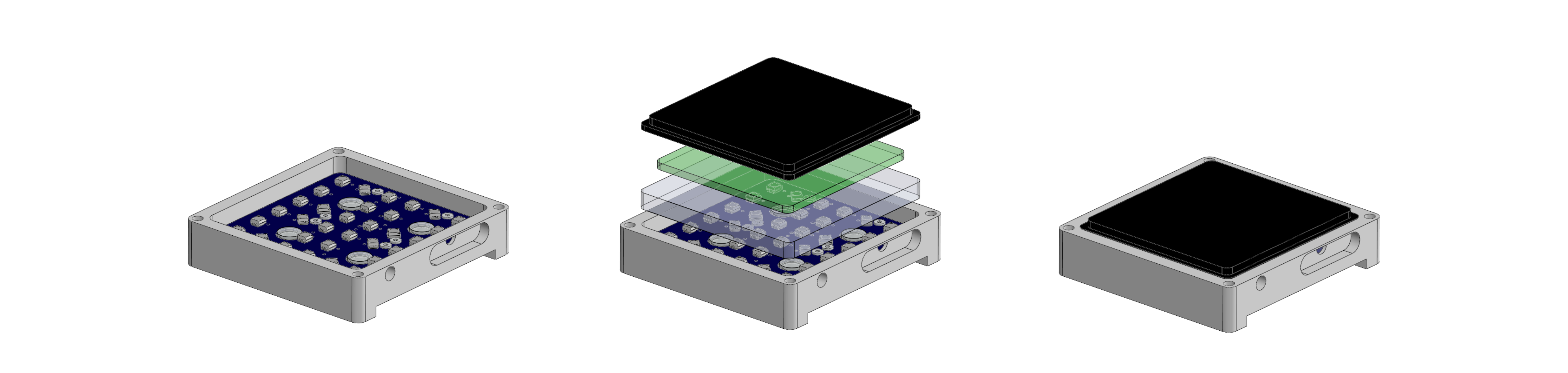}
\caption{The cameras and an LED board are fixed to a base structure. Three silicone layers are directly poured onto the LED board and the camera lenses: A stiff transparent layer, the particle layer and a black protection layer.}
\label{fig:sensor_structure}
\end{figure}

Similarly to \cite{sferrazza_sensors}, three different silicone layers are cast onto the camera array, as shown in Fig. \ref{fig:sensor_structure}. From the bottom, the first layer is relatively stiff and adds the distance between the camera and the particles, which is necessary to improve the focus. This layer also provides additional protection for the hardware and ensures light diffusion. The second layer is the softest and contains the particles tracked by the cameras. Finally, the third layer (stiffer than the second) is cast with a black color and protects the sensor from external light sources and material damage. The same materials, mixing ratios and curing protocol as in \cite{fem_access} were used, for a resulting sensing surface of 49$\times$51 mm.
Each embedded camera is controlled by a separate, relatively inexpensive single-board computer (Raspberry Pi 3 model B+). These boards communicate with a System-on-Module (NVIDIA Jetson Nano Developer Kit), which is equipped with a 64-bit quad-core Arm Cortex-A57 CPU alongside a Maxwell GPU with 128 CUDA cores. The communication is handled by a Gigabit Ethernet switch (ANDDEAR QZ001), which enables the Jetson Nano to receive the four image streams. The Jetson Nano provides a clock source to the Raspberry Pi boards, which are synchronized through the Networking Time Protocol (NTP), to ensure contemporaneous image streams. Note that the Raspberry Pi boards and the Ethernet switch may be replaced by compact, commercially available multi-camera adapter boards for the Jetson Nano. However, drivers for these adapter boards are still under development or not easily accessible because of the relatively recent release of the Jetson Nano. This aspect has not been further investigated for the purpose of this work.

\subsection{Dimensioning analysis}
\begin{figure}[!htb]
   \fontsize{7pt}{11pt}\selectfont
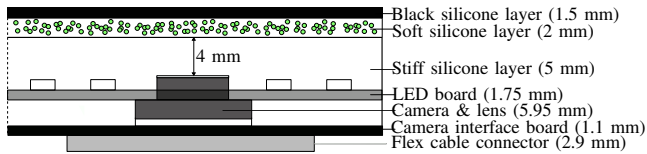
  \caption{A side-view schematic of the sensor's structure around one of the cameras is shown in this figure. Note that the overall thickness is determined by the two upper silicone layers, the distance between the lens and the particle layer, the camera and the camera interface board, including the connector. The LED board does not contribute to the overall thickness, since it is placed around the camera lenses.}
    \label{fig:height}
\end{figure}
The design presented above exhibits an overall thickness of 17.45 mm, which is lower than most of the camera-based tactile sensors described in the literature. As an example, compared to \cite{gelslim}, the sensor described here is slightly thinner and does not use mirrors, while covering a surface more than six times larger. In the following, some guidelines for further reducing the thickness of the sensor are detailed:
\begin{enumerate}
\item The commercial cameras employed in this work mount a flex cable connector at the bottom of their interface board, as shown in Fig. \ref{fig:height}. A custom camera interface board, with a connector placed in the space between the interface board itself and the LED board, may reduce the thickness by 2.9 mm, leading to an overall thickness of 14.55 mm.

\item A custom camera interface board may also be placed in a different position, farther from the cameras, depending on the application. Removing the camera boards and their connectors from below the cameras would result in an overall thickness of 13.45 mm. 

\item In the current design the interface boards are placed adjacent to each other in the same plane below the cameras. In order to cover a continuous surface, this requires that each camera covers a FOV of at least the size of an interface board. Moving the interface boards (as pointed out in the previous point) may additionally facilitate a closer placement of the cameras. As a consequence, this would make it possible to further reduce the distance between the lenses and the particles, while retaining a continuous surface coverage. Moreover, in this work the fisheye lenses were chosen among the commercially available solutions with a straightforward implementation. A tailored design with an accurate trade-off between the focal distance and the FOV may further reduce the overall thickness. Assuming an ideal pinhole camera model\footnote{A derivation of this fact in a simplified scenario is summarized in the online appendix \cite{online_appendix}.}, the thickness is mainly limited by the size of the image sensor. Modern image sensors with a thickness of about 0.3 mm are commercially available. The smallest commodity camera module\footnote{\url{https://www.ovt.com/sensors/OVM6948}} inclusive of a lens has a thickness of 1.158 mm, with the possibility of focusing a surface placed at a distance of 3 mm. Such a design may already result in a tactile sensor thickness of about \mbox{5 mm}.
\end{enumerate}

\section{METHOD} \label{sec:method}
In the following section the learning architecture is presented. First, the collection of the ground truth data is explained, then the details regarding the neural network are outlined.
\subsection{Data collection}
A dataset is collected following the strategy presented in \cite{fem_access}. Automated indentations are performed using a precision milling machine  (Fehlmann  PICOMAX  56  TOP)  with 3-axis  computer  numerical  control (CNC). On an evenly spaced grid, a spherically-ended cylindrical indenter with a diameter of 10 mm is pressed onto the sensor surface at different depths up to 1.5 mm. The same procedure is simulated with a finite element model in Abaqus/Standard \cite{abaqus_manual}, to assign ground truth labels to the images, representing the contact force distribution applied to the sensor's surface. In this regard, the surface is discretized into 650 bins of equal area. The procedure described in \cite{fem_access} provides the force applied to these bins, based on the FEM simulations. For each bin, three force components $F_x$, $F_y$, $F_z$ are provided, where $x$ and $y$ are aligned along the two horizontal sides of the sensor's surface and centered at one of the corners, and $z$ is the vertical axis, directed from the camera towards the sensor's surface. The resulting label vectors are assigned to the images from the four cameras for each indentation, and used in a supervised learning fashion, as described in the next subsection.

\subsection{Learning Architecture}
\label{sec:dnn}
\begin{figure*}
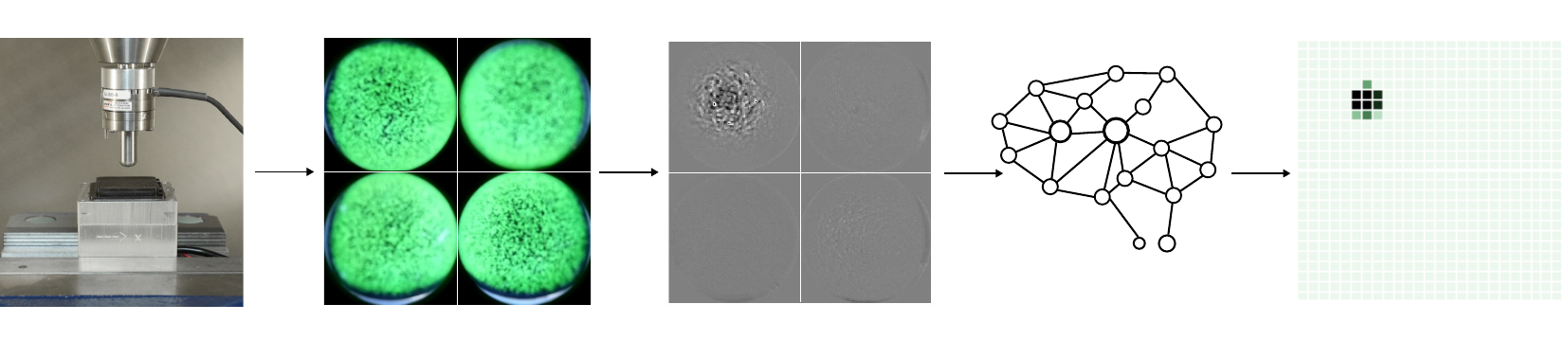
  \caption{The sensing pipeline is shown in this figure. An indentation produces a change in the particle pattern that is visible in the difference of the pixel intensities (central image) between the current frame and a frame taken at rest. The DNN predicts the three-dimensional contact force distribution applied during the indentation. The last figure on the right shows a color visualization of the resulting $F_z$ for each of the surface bins.}
  \label{fig:inference}
\end{figure*}
The prediction of the discretized force distribution applied to the surface of the tactile sensor is a multiple multivariate regression problem. This problem is tackled by training an end-to-end DNN that maps the images from the four cameras to the outputs of the network, that is, three force components for each of the 650 surface bins.
\begin{figure}[h]
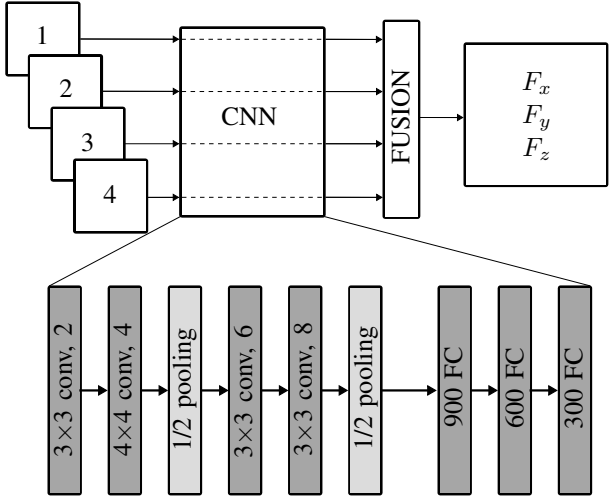
  \caption{This figure shows the architecture of the network. Each difference image is separately fed through the same CNN, and the outputs are then combined via a fusion layer. For ease of visualization, some abbreviations have been introduced in the block diagram above. In this regard, the label ``3$\times$3 conv, 2" refers to a two-channel convolutional layer with a 3$\times$3 kernel, while ``1/2 pooling" indicates a max pooling layer, which subsamples the input to half of its original size. ``900 FC" refers to a fully connected layer with 900 units.}
  \label{fig:architecture}
\end{figure}
The intensity difference images between the current frames and the respective four images taken in the undeformed surface state serve as the input to the network. An example of the resulting pipeline is shown in Fig. \ref{fig:inference}.

To decouple the detection of features that are independent of the cameras' placement, the difference images from the four cameras are fed independently through a convolutional neural network (CNN). Only after this intermediate network, a fusion layer with a linear activation function combines the four different output tensors and predicts the three-dimensional discretized force distribution. Both the overall network layout and the dimensions of the different layers are shown in Fig. \ref{fig:architecture}.

The CNN takes a difference image of size 128$\times$128 as an input. Batch normalization showed a large decrease in the network training times and is used for all convolutional layers, together with rectified linear unit activation functions. A dropout rate of 0.1 is used on the fully connected (FC) layers to prevent overfitting to the training data, together with sigmoid activation functions. The root mean squared error (RMSE) is used as a loss function for the Adam optimizer \cite{adam_optimizer} to train the model. 30\% of the dataset was put aside for evaluation purposes.

The fact that the CNN is shared between the four cameras leads to a considerable reduction in the memory consumption, especially in view of the possible extension to larger surfaces with a higher number of cameras. Moreover, this generally leads to a smaller network size, in contrast to feeding the concatenation of the four images through a larger architecture. Smaller architectures tend to require less training data and exhibit shorter training times. Finally, the architecture presented here shows modularity features. These are discussed in Section \ref{sec:results}.

\section{RESULTS} \label{sec:results}
In this section, the performance of the multi-camera tactile sensor is evaluated. In the first part of the section, the evaluation of the learning architecture is presented. In the second part, an experiment is performed by modifying the neural network and the training procedure in order to test the modularity of the approach.

\subsection{Sensor Performance}
\label{sec:original}
\begin{figure}[htb]
\centering
\hspace*{-15pt}
\input{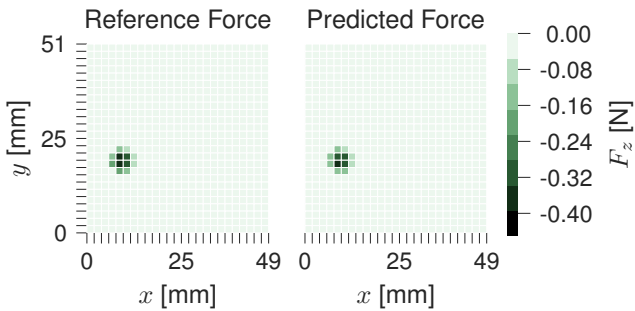}
\caption{The figure shows the component $F_z$ of each surface bin for an example indentation in the test set. On the left, the ground truth force distribution is shown, and on the right, the distribution predicted by the neural network.} 
\label{fig:pred}
\end{figure}

The DNN presented in Section \ref{sec:dnn} is trained on 70\% of the full dataset. 10\% of the samples are used as a validation set to apply early stopping during training. The remaining 20\% is left aside for evaluation. The resulting RMSE on the force distribution is 0.00060 N, 0.00059 N, 0.0019 N for $F_x$, $F_y$, $F_z$, respectively, while the resulting RMSE on the total applied force (sum of the forces of all surface bins) is 0.0019 N, 0.0016 N, 0.0571 N for $F_x$, $F_y$, $F_z$, respectively. Note that the dataset is generated from vertical indentations, with the resulting total forces in $z$-direction up to 3 N, and considerably smaller total forces in the horizontal direction on most of the sensor surface. Fig. \ref{fig:pred} shows an example prediction of the contact force distribution in $z$-direction and the corresponding ground truth.
The model inference runs on the Jetson Nano at 86 Hz, which makes the frame capturing (40 frames per second) on the Raspberry Pi boards the bottleneck of the sensing speed. As a result, the sensing pipeline runs at a maximal prediction speed of 40 Hz. Furthermore, the fact that the four simultaneous images are fed independently through the CNN enables the detection of multiple contact points, when each camera fully captures up to one of the distinct contact patches, even if the model has only been trained with single indentations. This makes it possible to detect up to four distinct contact patches, as shown in the video attached to this submission.\\
%

\subsection{Sensor Modularity}
\begin{figure}[!htb]
	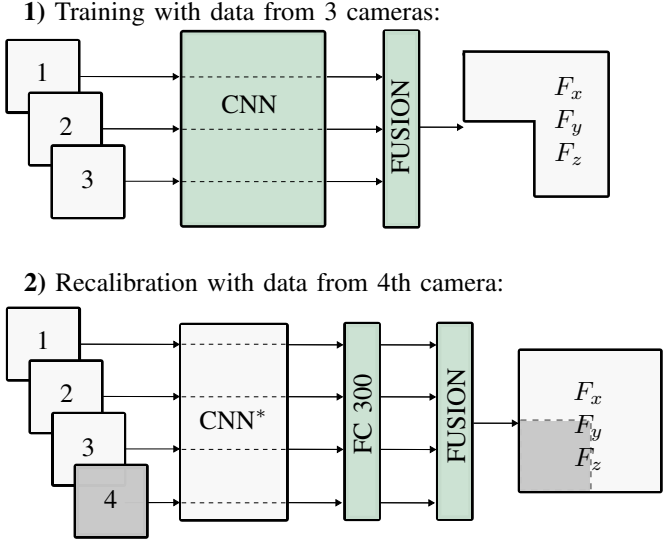
	\caption{The model is first trained with the data from three cameras, and then extended to four cameras and recalibrated with the full dataset. In the first plot the fusion layer contains $900$ units, mapping the outputs corresponding to the surface covering only three cameras. In the second plot, the size of the fusion layer is increased to $1200$ units, to cover the entire surface.}
	\label{fig:mod}
\end{figure}

To evaluate the modularity of the approach, a first model is trained using only images and labels from three cameras. In a second step, the sensor is recalibrated with the training data from all four cameras. The procedure is schematically shown in Fig. \ref{fig:mod}. For the calibration step, the majority of the DNN parameters are frozen, and only the last fully connected layer and the fusion layer are retrained. This serves the purpose of reducing the training times and the data requirements.

\begin{figure*}
        \centering
        \begin{subfigure}[b]{0.475\textwidth}
            \centering
           \resizebox{0.8\columnwidth}{!}{\input{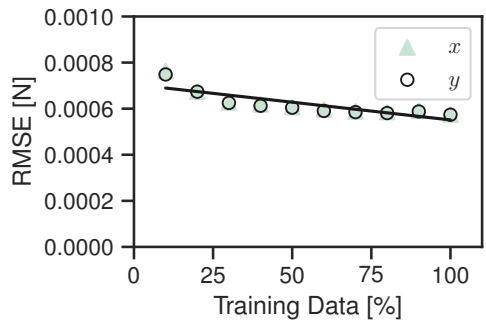}}
            \caption%
            {{\small RMSE on the force distribution ($x$ and $y$ axes)}} 
        \end{subfigure}
        \hfill
        \begin{subfigure}[b]{0.475\textwidth}  
            \centering 
            \resizebox{0.8\columnwidth}{!}{\input{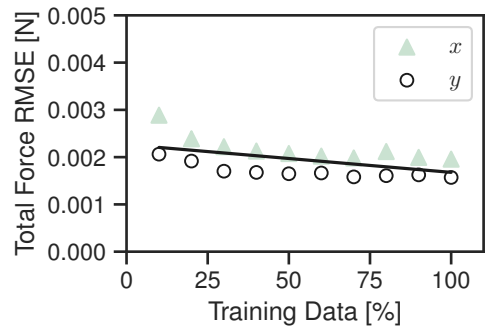}}
            \caption{{\small RMSE on the total force ($x$ and $y$ axes)}}    
        \end{subfigure}
        \vskip\baselineskip
        \begin{subfigure}[b]{0.475\textwidth}   
            \centering 
            \resizebox{0.8\columnwidth}{!}{\input{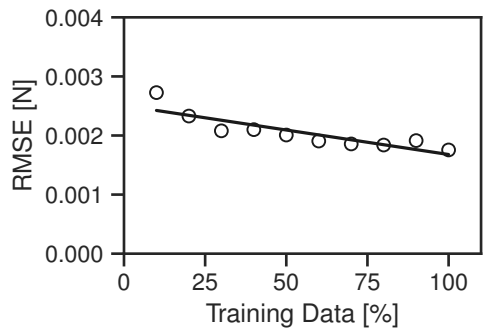}}
            \caption{{\small RMSE on the force distribution ($z$ axis)}}    
        \end{subfigure}
        \quad
        \begin{subfigure}[b]{0.475\textwidth}   
            \centering 
            \resizebox{0.8\columnwidth}{!}{\input{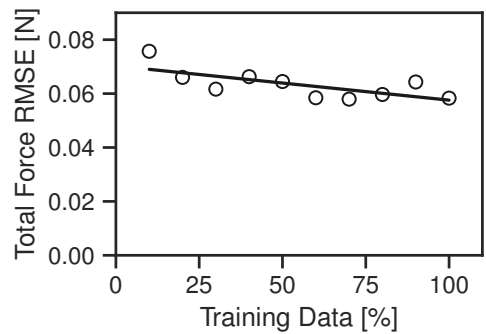}}
            \caption{{\small RMSE on the total force ($z$ axis)}}    
        \end{subfigure}
        \caption{After the neural network is trained on the data from three cameras, it is recalibrated for the fourth camera with a portion of the training data from all the cameras. The resulting errors are shown above, as a function of the varying percentage of the full dataset used for training, together with a least-squares trend line. } 
        \label{fig:data}
            \end{figure*}

As shown in Fig. \ref{fig:data}, the recalibrated network shows comparable performance to the model trained on the whole data in Section \ref{sec:original}. Moreover, the performance is retained using an even smaller portion of training data. The plots show the different error metrics as a function of the percentage of data used for training. Retraining the last two layers takes approximately 1.5 hours on the employed GPU (Nvidia TITAN X Pascal), as opposed to over 10 hours training for the whole model on the full dataset. This experiment shows promising results towards the possibility of training the most resource (both time and data) consuming part of the network on a subset of the surface, therefore reducing the data collection and the training times on the rest of the surface. This also opens the opportunity of replacing a defective camera (although not currently possible in the experimental prototype presented) on a large-scale skin, without the need to retrain the entire network.\\

\section{CONCLUSION} \label{sec:conclusion}
In this paper, a multi-camera tactile sensing approach has been presented, making use of an end-to-end DNN to predict the force distribution on the sensor surface. The sensor thickness has been reduced, while at the same time the sensing surface has been extended using multiple cameras arranged in an array. A relatively inexpensive sensor design has been suggested, using Raspberry Pi cameras to capture synchronized images and a Jetson Nano Developer Kit for the model inference. 

A neural network has been presented to reconstruct the contact force distribution. The modular architecture proposed here is applicable to optical tactile sensors with a larger numbers of cameras, such as vision-based robotic skins. It has been shown how the network can be efficiently recalibrated when the number of cameras is increased, without retraining any of the convolutional layers. The sensing pipeline presented here runs on an embedded computer provided with a GPU at 40 Hz. On the test dataset employed, the architecture has shown an RMSE of \mbox{0.0571 N} on the total forces in the vertical direction that were collected up to 3 N.

The procedure proposed here to attach the lenses to the camera frames does not yield a very accurate focus of the images (see Fig. \ref{fig:inference}) and needs further investigation. Future work will also include the extension of this approach to various directions and shapes of the indentations, as well as a quantitative scalability analysis in the case of an increasing number of cameras.



\section*{ACKNOWLEDGMENT}
The  authors  would  like  to  thank  Michael  Egli  and Matthias M{\"u}ller for their support in the sensor manufacture.


\bibliographystyle{IEEEtran}
\bibliography{references}

\end{document}